\documentclass[journal]{IEEEtran}

\usepackage{caption}
\usepackage{graphicx}
\usepackage{amsmath}
\usepackage{booktabs}
\usepackage{setspace}
\usepackage{cite}
\usepackage[square, comma, sort&compress, numbers]{natbib}
\usepackage{cases}
\usepackage{pifont}
\usepackage{array}
\usepackage{subfigure}

\ifCLASSINFOpdf
\else
\fi
\begin{document}
	
	\title{Generating Text Sequence Images for Recognition}
	% A method for Generating Text Sequence Images
	% Generating Text Sequence Images with Adversarial Networks
	% Generating Text Sequence Images without Quantitive Limits
	% Generating Text Sequence Images for Recognition
	
	\author{Yanxiang~Gong\textsuperscript{1},
		Linjie~Deng\textsuperscript{1},
		Zheng~Ma
		and~Mei~Xie\textsuperscript{*}% <-this % stops a space
		\thanks{The authors are with the School of Information and Communication Engineering, University of Electronic Science and Technology of China, Chengdu 611731, Sichuan, China.}
		\thanks{1 These authors contributed equally to this work.}
		\thanks{*Mei Xie is the corresponding author(e-mail:mxie@uestc.edu.cn).}}% <-this % stops a space}% <-this % stops a space
		%\thanks{J. Doe and J. Doe are with Anonymous University.}% <-this % stops a space
		%\thanks{Manuscript received April 19, 2005; revised August 26, 2015.}}

	% The paper headers
	%\markboth{Journal of \LaTeX\ Class Files,~Vol.~14, No.~8, August~2015}%
	%{Shell \MakeLowercase{\textit{et al.}}: Bare Demo of IEEEtran.cls for IEEE Journals}
	
	% make the title area
	\maketitle
	
	\begin{abstract}
Recently, methods based on deep learning have dominated the field of text recognition. With a large number of training data, most of them can achieve the state-of-the-art performances. However, it is hard to harvest and label sufficient text sequence images from the real scenes. To mitigate this issue, several methods to synthesize text sequence images were proposed, yet they usually need complicated preceding or follow-up steps. In this work, we present a method which is able to generate infinite training data without any auxiliary pre/post-process. We tackle the generation task as an image-to-image translation one and utilize conditional adversarial networks to produce realistic text sequence images in the light of the semantic ones. Some evaluation metrics are involved to assess our method and the results demonstrate that the caliber of the data is satisfactory. The code and dataset will be publicly available soon. 
	\end{abstract}
	% Note that keywords are not normally used for peerreview papers.
	
	\IEEEpeerreviewmaketitle
	
	\section{Introduction}
	
	\IEEEPARstart{T}{ext} recognition plays an important role in the field of computer vision. With the advent of deep learning, text recognition methods have made great progresses\cite{Jaderberg14c,lee2016recursive,crnn,fan2018edit}. But they cannot achieve a satisfactory performance for insufficient training data which causes over-fitting problems. Owing that to collect and label real text images is a time-consuming work, methods to synthesize text images were proposed in order to alleviate the deficit in training data. The method put forward by Jaderberg \textsl{et al.}\cite{Jaderberg14c,Jaderberg14d} is based on a font catalogue. Coloring and projective distortion are applied on word images synthesized by font, border and shadow rendering, and then the processed images are added to background scene images with some noises. Gupta \textsl{et al.}\cite{Gupta16} proposed to apply the semantic segmentation on the scene image at first. Then the processed word images are pasted on a contiguous region of it, which guarantees that the word will not appear on objects of different distances. Based on \cite{Gupta16}, Zhan \textsl{et al.}\cite{verisimilar} presents a method which realizes semantic coherent synthesis. By leveraging the semantic annotations of objects and image regions created in the prior semantic segmentation research, semantic coherence between the text and the background has been reached while synthesizing text images. These methods are effective, but usually need complicated preceding or follow-up steps such as collecting background images, coloring the words and adding noises for improving the robustness, which requires more manual engineering.
	
	In this paper, we propose a method based on Generative Adversarial Networks(GANs)\cite{goodfellow2014generative} which can generate infinite realistic text sequence images without any extra pre/post-process. The inspiration comes from the procedure of drawing pictures. While painting something in the real world, generally we will sketch the contours of it at first, and then use pigments to color the draft to finish the drawing. Following this procedure, we cope with the generation task from a new perspective as an image-to-image translation one, and utilize conditional adversarial networks to yield text sequence images on the basis of semantic ones. This work is based on a modified conditional Generative Adversarial Networks\cite{cgan} model named pix2pix\cite{pix2pix} which aims to translate semantic images to realistic ones. Some evaluation metrics will be utilized to assess our method and confirm the effectiveness. There are two main contributions in this work:
	
	1.Unlike previous approaches, our method needs no extra preceding and follow-up step for generation. Besides, infinite images can be produced without any redundant operations.
	
	2.The data generated by our method achieves a satisfactory performance on various evaluation metrics, and the code and dataset will be publicly available soon.
	
	\begin{figure}
		\centering
		\subfigure[]{\includegraphics[width=2.8in]{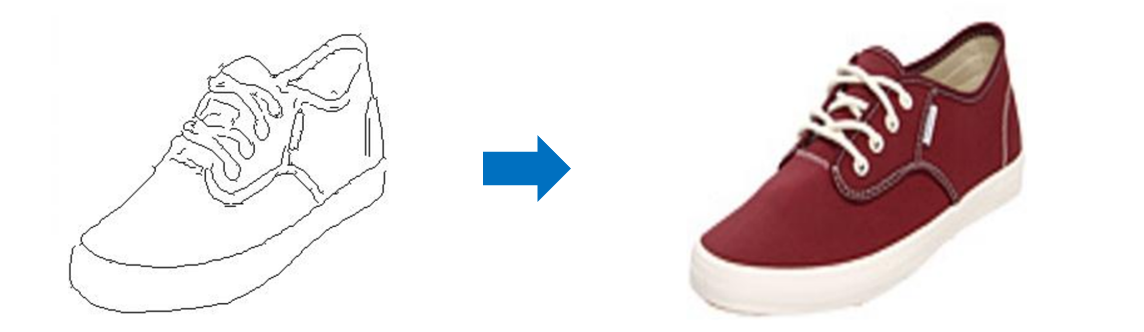}}
		\subfigure[]{\includegraphics[width=2.8in]{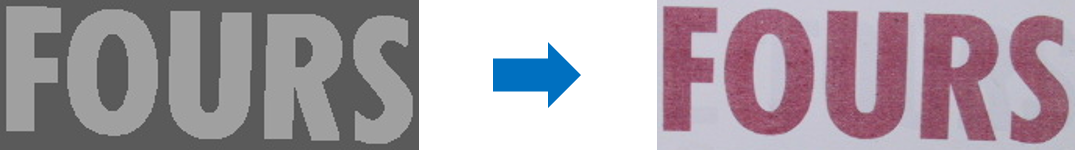}}
		\caption{Image translation from the sketch to the real image by the pix2pix network(a) and translation from the semantic text image to the scene text image by our method(b). The images on the left side are semantic ones, and on the right side are generated ones.}
		\label{pipelines}
	\end{figure}
	
	\begin{figure*}
		\centering
		\subfigure[]{\includegraphics[width=2.8in]{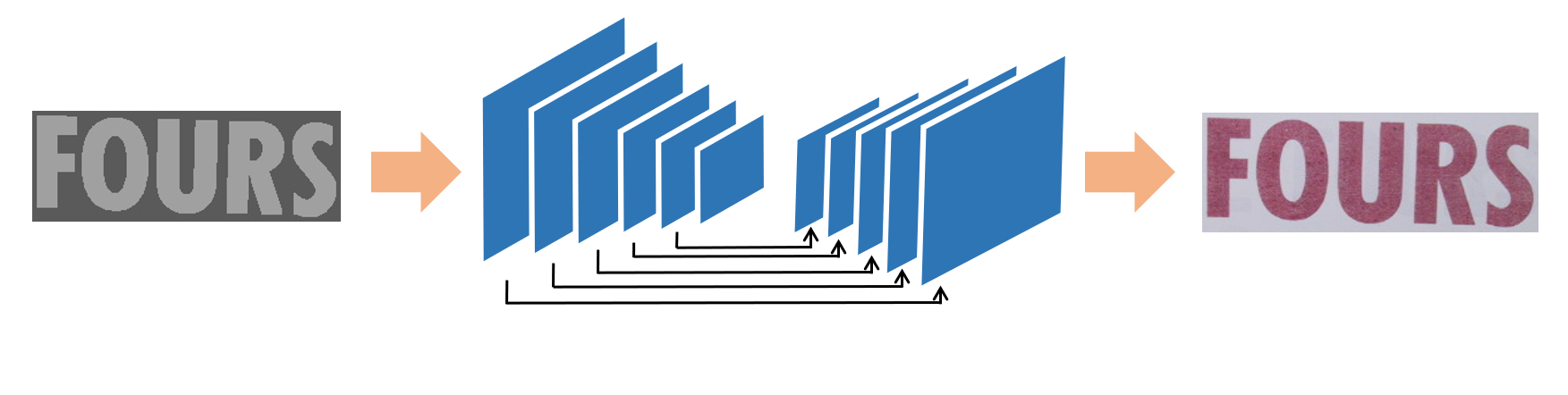}}
		\subfigure[]{\includegraphics[width=2.8in]{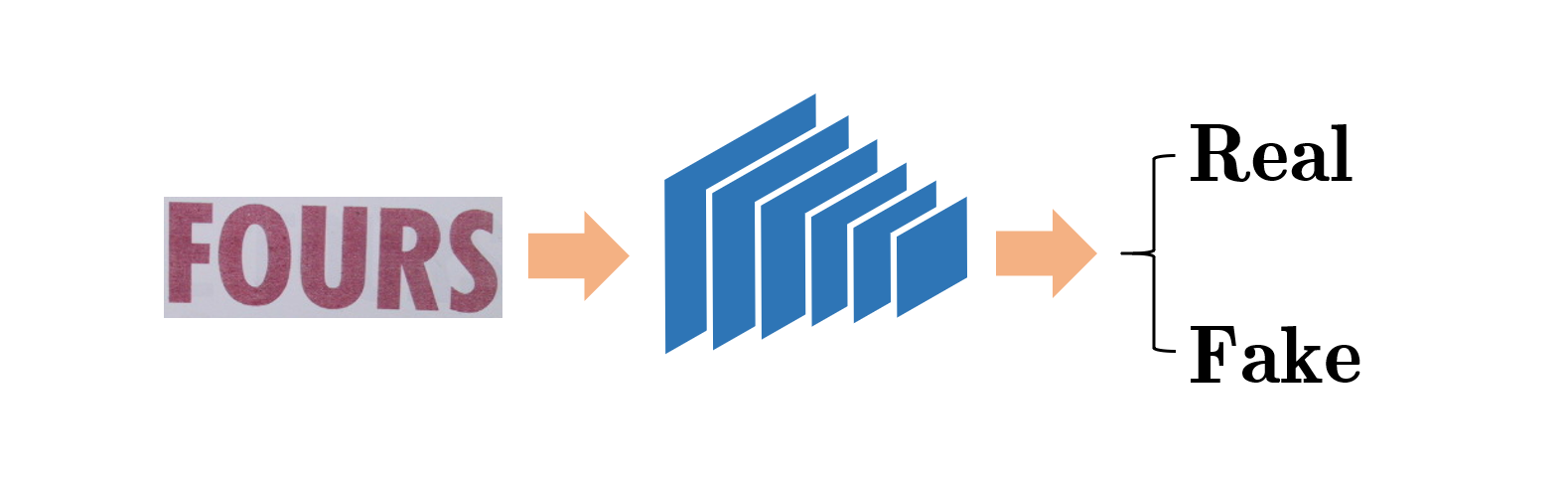}}
		\subfigure[]{\includegraphics[width=5.6in]{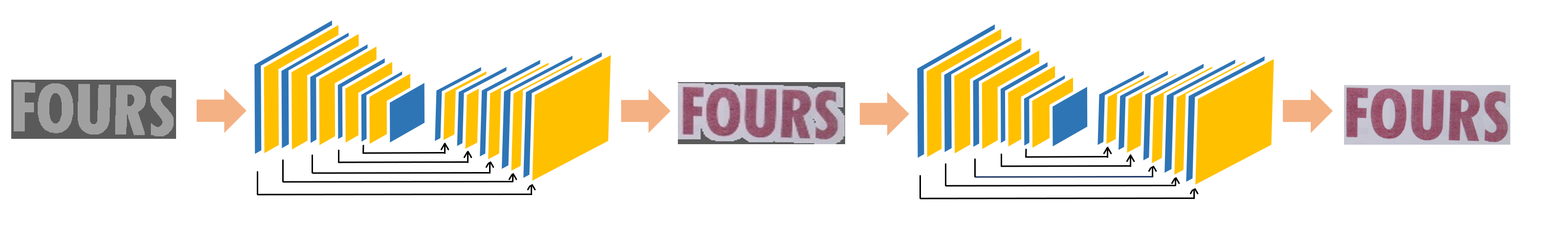}}
		\subfigure[]{\includegraphics[width=4in]{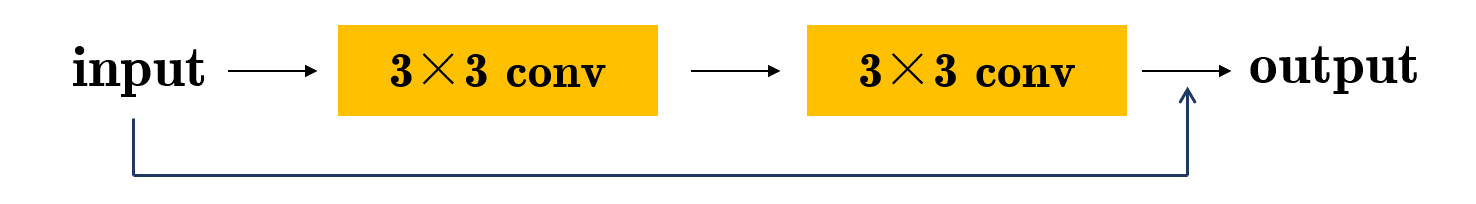}}
		\caption{The architecture of networks: (a) The generator of the initial pix2pix; (b) The discriminator of the initial pix2pix; (c) The generator of our network. The discriminator of our network is the same as the one in the initial pix2pix; (d) The architecture of the residual block. The blue and yellow rectangles in (a), (b) and (c) represent the feature maps generated by convolutional module and residual block respectively. The black lines represent the skip connections.}
		\label{networks}
	\end{figure*}
	
	\section{Method}
	
	Recently, works based on Generative Adversarial Networks(GANs)\cite{goodfellow2014generative, cgan, pix2pix} has made great achievements in the field of image generation. The initial GANs\cite{goodfellow2014generative} are models which learn a mapping from random noise vector $z$ to output image $y$, $G:z\rightarrow y$. By contrast, the subsequently proposed condition Generative Adversarial Networks(cGAN)\cite{cgan} are models that learn a mapping from an image $x$ and random noise vector $z$ to output image $y$, $G:(x,z)\rightarrow y$. We will introduce the details of our methods in the following parts.
	
	%The pipelines of pix2pix\cite{pix2pix} and our work are shown in Fig. 1.
	
	\subsection{Network Architectures}
	First let us recall the architecture of pix2pix network\cite{pix2pix}. It uses modules of the form Convolution-BatchNorm-ReLU\cite{ioffe2015batch} in both the generator and the discriminator. And inspired by U-Net\cite{unet}, some skip connections are added to an encoder-decoder network\cite{hinton2006reducing} as the generator. The architectures of the generator and the discriminator of the pix2pix model\cite{pix2pix} are shown respectively in Fig. 2(a) and Fig. 2(b). The objective can be expressed as
	\begin{equation}
	\min \limits_{G}\max\limits_{D}{L_{cGAN}(G,D)+\lambda L_{L1}(G)}\label{demo4}
	\end{equation}
	where $L_{cGAN}(G,D)$ represents the objective of condition GANs and $L_{L1}(G)$ represents the L1 distance. The parameter $\lambda$ is set to 100 in the original work. We make some adaptations on the network in order to make it more appropriate for text images. Each component will be listed in the following part, and ablation experiments which can confirm that they are effective will be described in next chapter. The pipelines of pix2pix\cite{pix2pix} and our work are shown in Fig. 1.
	%Ablation experiments which can confirm that the adaptations are effective will be described in next chapter. The architecture of the adapted generator is shown in Fig. 2(c) and each of the adaptations will be introduced in detail below.
	
	\subsubsection{Cascaded generators}
	Enlightened by StackGAN\cite{stackgan} who decomposes the text-to-image generative process into two stages, where the first GANs model aims to generate images with a small size and the second one tries to improve the resolution, we cascade two generators to make the generator to possess its own focus. The first generator is designed to generate the text area and its surroundings which are defined as the foreground area obtained through dilation operation on the masks of the text sequence images. The second generator aims to supplement the background area to produce realistic scene images. The architecture of the two generators are the same, while they are optimized respectively. As the generators do not need to generate too many areas, they can concentrate on their own work. Restrict by the hardware, only two generators are utilized though more generators are available. The procedure is depicted in Fig.2(c).
	
	%The mask of the synthesized image is dilated to get the surrounding area at first, as shown in Fig. 3(a). Then we synthesize the foreground images whose text area and its surroundings is the same as real images and the other area is the same as semantic images. The generator is supposed to translate the semantic text images to foreground images. 
	% \begin{figure}
	%	\centering
	%	\includegraphics[width=.35\textwidth]{images/final/dilation.png}
	%	\caption{The processes to get surroundings of the text area(a) and synthesizing foreground images(b)}
	%	\label{img3}
	%\end{figure}
	
	\begin{table*}[!t]
		\renewcommand{\arraystretch}{1.3}
		\caption{\label{table1}Ablations for proposed component. We evaluate the effectiveness of each component with some metrics, like Inception score, FID score and the performances of recognition on some public benchmarks.}
		\centering
		\begin{tabular}{|p{1.5cm}<{\centering}|p{1.5cm}<{\centering}|c|c|c|p{1.7cm}<{\centering}|p{1.7cm}<{\centering}|p{1.2cm}<{\centering}|}
			\hline
			Cascaded Network & Residual Blocks & PReLU  & Inception score & FID score & ICDAR2013 & ICDAR2015 & IIIT 5K \\
			\hline
			&            &                 & 3.002$\pm$0.257          & 51.05          & 70.2      & 43.9     & 68.4        \\
			\hline
			&            &   \ding{52}     & 2.877$\pm$0.198          & 48.27          & 72.4      & 45.9     & 71.1       \\
			\hline
			& \ding{52}  &                 & 2.998$\pm$0.203          & 45.75          & 75.8      & 47.9     & 73.7         \\
			\hline
			\ding{52}   &            &                 & 3.086$\pm$0.216          & 51.10          & 75.2      & 47.1 & 75.3         \\
			\hline
			\ding{52}   & \ding{52}  & \ding{52}       & \textbf{2.821$\pm$0.174} & \textbf{39.44} & \textbf{78.7} & \textbf{51.0} & \textbf{78.0}\\
			\hline
		\end{tabular}
	\end{table*}
	
	\begin{table*}[!t]
		\renewcommand{\arraystretch}{1.3}
		\caption{\label{table1}Comparison experiments with other datasets. The evaluation metrics are the same with Table I. We also involve the real images which are utilized to train our GANs model.  }
		\centering
		\begin{tabular}{|c|c|c|c|p{1.7cm}<{\centering}|p{1.7cm}<{\centering}|p{1.2cm}<{\centering}|}
			\hline
			Dataset          & Image number & Inception score            & FID score        & ICDAR2013 & ICDAR2015& IIIT 5K \\
			\hline
			Real images    & 6k           & 3.363$\pm$0.404            & \textbf{25.93}   & 12.0       &  1.9  & 5.3      \\
			\hline
			Gupta \textsl{et al.}\cite{Gupta16}        
			& 8M           &  4.115$\pm$0.202          &  66.16      & 78.3       &  46.7  & 75.5   \\
			\hline
			Jaderberg \textsl{et al.}\cite{Jaderberg14c,Jaderberg14d}        
			& 8M           & \textbf{2.747$\pm$0.170}   & 51.68          & 78.1         &  \textbf{52.1}  & 77.6   \\
			\hline
			Ours           & 8M           &  2.821$\pm$0.174          &39.44            &\textbf{78.7}  &   51.0  & \textbf{78.0}  \\
			\hline
			Gupta \textsl{et al.}\cite{Gupta16}(colored)        
			& 8M           &  4.848$\pm$0.323       &  72.98           & 78.7 & 48.5 & 74.9 \\
			\hline
			Ours(colored)  & 8M      & 3.163$\pm$0.175  &  45.46    & 78.6 & 48.7 & 72.4\\
			\hline
		\end{tabular}
	\end{table*}
	
	\subsubsection{Residual Blocks}
	Residual Networks(ResNets)\cite{resnet} solves the degradation problem in the process of training a deeper network through employing some residual blocks. The mapping of a residual block can be expressed as
	\begin{equation}
		H(x)=F(x) + x
	\end{equation}
	which let the layers fit a residual mapping rather than a desired underlying mapping to mitigate the vanishing gradient problem. In order that our model can be optimized better, some residual blocks are added to the generator. As shown in Fig 3(c), after each convolutional layer except the last one of the encoder, a residual block with two $3\times3$ convolutional layers which is depicted in Fig. 2(d) will be added.
	\subsubsection{Activation Function}
	For better abilities to extract the features, we change the activation function of the encoder from leaky Rectified Linear Unit(leaky ReLU) to Parametric Rectified Linear Unit (PReLU)\cite{prelu}.
	
	\begin{subnumcases}
	{PReLU(x)=}
	ax, &$x\le0$,\\
	x, &$x>0$.
	\end{subnumcases}
	
	The parameter $a$ are set to 0.25 at first, and it will be updated automatically while training. In contrast, the parameter of leaky ReLU should be set up by ourselves, while to search for the best fitted parameter will take lots of time without satisfactory effects. Because there are only a few parameters added to the network, the computation and risks of over-fitting will not increase too much.

	\subsection{Synthesizing Semantic Images}
	Gupta \textsl{et al.}\cite{Gupta16} proposed a method to synthesize text sequence images through morphology ways, which gives us inspiration about synthesizing semantic images. Taking this approach as basis, first we acquire suitable text samples from Newsgroup20 dataset\cite{newsgroups} in words, lines and paragraphs. Then the text sample is rendered with a randomly selected font
	and transformed randomly. Finally the text is blended into a black background image using Poisson image editing\cite{possion}.
	
	\section{Experiments}
	In the following part, we will describe the implementation details of our method. We also utilize some evaluation metrics and run a number of ablations to analyze the effectiveness of the proposed component.
	\subsection{Training}
	In training stage, we collect some data from ICDAR 2013 training dataset\cite{icdar2013} which contains 229 images and KAIST scene text database\cite{kaist} which contains 1,498 images. It is worth noting that no testing dataset is involved into the training process. There are totally 6,715 word images while we discard those who only contain punctuations and those whose height is longer than the width, and we relabel them for better adaptability of our model. In the training stage, the optimizer of the network is Adam\cite{adam}, the batch size is set to 64, and the learning rate is 0.0002. All images will be resized to $128\times64$. The network is trained for 200 epochs which consumes about 2 hours. The proposed method is implemented by PyTorch\cite{pytorch}. All experiments are carried out on a standard PC with Intel i7-8700 CPU and a single NVIDIA TITAN Xp GPU. 
	
	\begin{figure*}
	\centering
	\subfigure[]{\includegraphics[width=2.3in]{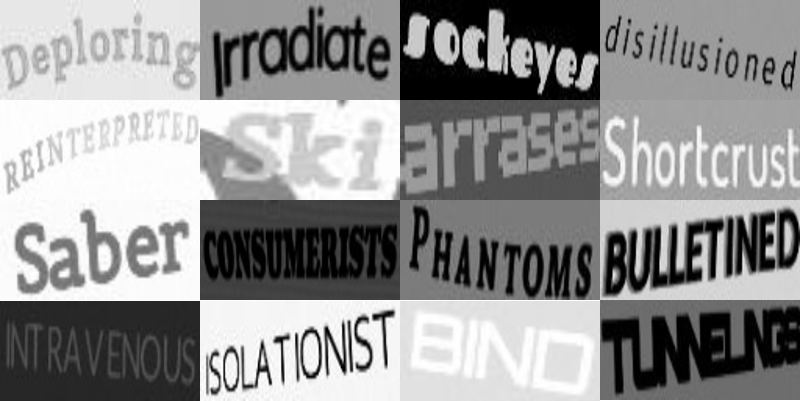}}
	\subfigure[]{\includegraphics[width=2.3in]{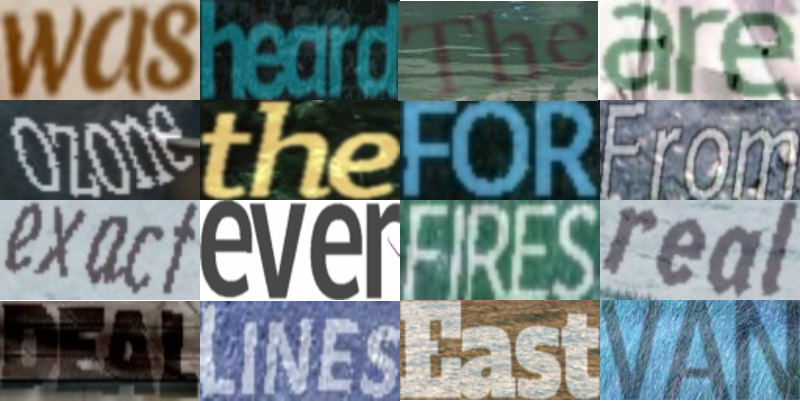}}
	\subfigure[]{\includegraphics[width=2.3in]{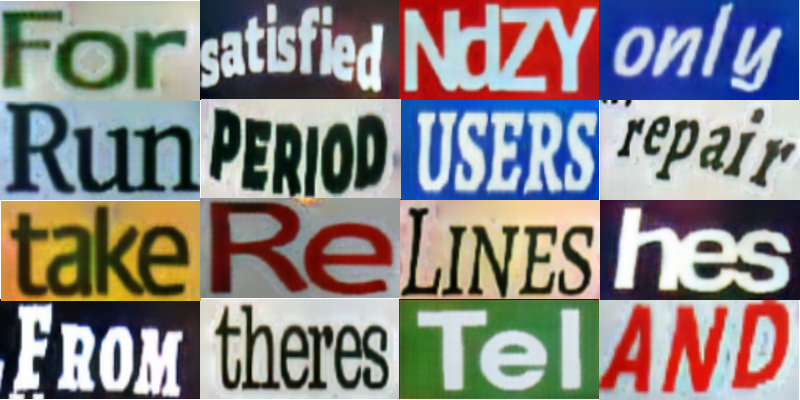}}
	\caption{Images produced by Jaderberg \textsl{et al.}(a), Gupta \textsl{et al.}(b) and our method(c). Best viewed in color.}
	\label{img6}
\end{figure*}
	
	\subsection{Evaluation Metrics}
	
	\subsubsection{Inception score}
	To calculate Inception score\cite{IS} of the generated dataset is a way to evaluate its quality. Images that contain meaningful objects should have a conditional label distribution $p(y|x)$ with low entropy. And the marginal $\int p(y|x = G(z))dz$ should have high entropy owing that we expect the model to generate varied images. 
	
	\subsubsection{FID score}
	Fr\'{e}chet Inception Distance(FID) score\cite{fid} is also an indicator of the performance of a model of GANs because it represents the similarity between two datasets. A lower FID score means two datasets are more similar with each other. As we expect the distribution of the generated images to be close to real ones, FID scores between the generated data and ICDAR 2013 testing dataset\cite{icdar2013} are utilized to evaluate our model.
	
	\subsubsection{Recognition Task}
	Actually, the initial intention is to generate training data for recognition models. A higher accuracy of a trained model will prove that the data has a better quality. Therefore, an end-to-end recognition network named CRNN(Convolutional Recurrent Neural Network)\cite{crnn} is applied to test our model. The text sequence images and the contents of them will be fed into the training stage. For fair comparisons, we generate 8M images and transform them to gray ones to match the data from Jaderberg \textsl{et al.}\cite{Jaderberg14c, Jaderberg14d}. In addition, we also test the model through using colored images. In the training process, the batch size is set to 64 and the learning rate is 0.00005 with the SGD optimizer. The network is trained for 3 epochs which consumes about 6 hours carried on our hardware. The trained models will be evaluated on some public benchmarks such as ICDAR 2013\cite{icdar2013}, ICDAR 2015\cite{icdar2015} and IIIT 5K\cite{IIIT5K}.%and Street View Text(SVT) dataset\cite{svt}. 
	% There is also a way to test the quality of the generated data through training a recognition model. The accuracy of the model will be higher if the data has a better quality. Therefore, Convolutional Recurrent Neural Network (CRNN)\cite{crnn}, which is an end-to-end recognition network, is applied to test our model. The inputs of the network are text sequence images, and it can yield the content of the text directly. We train CRNN networks\cite{crnn} with 8000k data which have been transformed to gray ones and test the trained model on ICDAR 2013 testing dataset\cite{icdar2013} and IIIT 5K Dataset\cite{IIIT5K}. 

	\subsection{Ablation Experiments}
	% There are two steps to evaluate our method. First is to compare with different adapted networks to confirm that our adaptations are effective. The initial pix2pix network and the networks using only one generator without cascading, removing the residual blocks and changing the activation function of the encoder back to leaky ReLU are involved. Second is comparisons with other datasets. The same number of images generated by Jaderberg \textsl{et al.}\cite{Jaderberg14c, Jaderberg14d}, Gupta \textsl{et al.}\cite{Gupta16} and our generated data are utilized. Some of the images of these datasets are shown in Fig. 3.
	
	%  
	
	First, we evaluate the effectiveness of the proposed components and compare them with our baseline models, which is extended straightly from the pix2pix\cite{pix2pix} framework. The comparison results are listed in Table I. From the table, we can observe that each component achieves a progress of the performance compared with the baseline model. We integrate them and get a further promotion of each evaluation metric, which demonstrate the proposed components are effective for the generation task.
	
	Second, we make some comparisons with other methods including Jaderberg \textsl{et al.}\cite{Jaderberg14c, Jaderberg14d} and Gupta \textsl{et al.}\cite{Gupta16}. The numbers of images generated by each method are the same. Some samples of each dataset are shown in Fig. 3. Specially, we involve the training data of our GANs model into the comparisons. The results are shown in Table II. Naturally, the training images of our GANs model achieve the best FID score because they are sampled from the same distribution of ICDAR 2013 testing dataset. But they cannot achieve a good performance for in the recognition task cause there is a huge over-fitting problem within only 6k training images. In the inception scores, Jaderberg \textsl{et al.} reaches the best owing that the images contains less background information. The colored images are not as good as gray ones because a colored background contains more contents. In the FID scores, our method get the first place in the three producing methods. In recognition task, we achieve the highest accuracy on ICDAR 2013 and IIIT 5K. On ICDAR 2015, our data cannot achieve the best accuracy. We consider that it is because ICDAR 2015 contains plenty of images whose text is vague or distorted, and our generated images are too clear. In addition, the colored images are also evaluated but there is no obvious promotion. We argue that the CRNN network\cite{crnn} is not sensitive about the color mode of inputs. Finally, it is worth noting that each of the recognition model are only trained with 8M data, but our method is able to generate infinite images without any extra process.

	\section{Conclusion and Future Work}
	We have proposed a method to generate realistic text sequence images for training recognition models. The method is able to produce infinite images with high quality, which exceeds general morphology methods. As more complicated networks can be used to synthesize high resolution images, in the future, our goal is to design an end-to-end system that can detect and recognize text in an image with high resolution while given a font catalogue and a lexicon finally.
	
	% Can use something like this to put references on a page
	% by themselves when using endfloat and the captionsoff option.
	\ifCLASSOPTIONcaptionsoff
	\newpage
	\fi
	
	% trigger a \newpage just before the given reference
	% number - used to balance the columns on the last page
	% adjust value as needed - may need to be readjusted if
	% the document is modified later
	%\IEEEtriggeratref{8}
	% The "triggered" command can be changed if desired:
	%\IEEEtriggercmd{\enlargethispage{-5in}}
	
	% references section
	
	% can use a bibliography generated by BibTeX as a .bbl file
	% BibTeX documentation can be easily obtained at:
	% http://mirror.ctan.org/biblio/bibtex/contrib/doc/
	% The IEEEtran BibTeX style support page is at:
	% http://www.michaelshell.org/tex/ieeetran/bibtex/
	%\bibliographystyle{IEEEtran}
	% argument is your BibTeX string definitions and bibliography database(s)
	%\bibliography{IEEEabrv,../bib/paper}
	%
	% <OR> manually copy in the resultant .bbl file
	% set second argument of \begin to the number of references
	% (used to reserve space for the reference number labels box)
	%\begin{thebibliography}{1}
	
	%\bibitem{IEEEhowto:kopka}
	%H.~Kopka and P.~W. Daly, \emph{A Guide to \LaTeX}, 3rd~ed.\hskip 1em plus
	%  0.5em minus 0.4em\relax Harlow, England: Addison-Wesley, 1999.
	
	%\end{thebibliography}
	
	%\bibliographystyle{ieeetr}
	%\bibliography{ref}
	% \cite{Jaderberg14c}
	%\section*{References}
	
	\bibliographystyle{IEEEtran}
	\bibliography{IEEEabrv,bare_jrnl}

	% biography section
	% 
	% If you have an EPS/PDF photo (graphicx package needed) extra braces are
	% needed around the contents of the optional argument to biography to prevent
	% the LaTeX parser from getting confused when it sees the complicated
	% \includegraphics command within an optional argument. (You could create
	% your own custom macro containing the \includegraphics command to make things
	% simpler here.)
	%\begin{IEEEbiography}[{\includegraphics[width=1in,height=1.25in,clip,keepaspectratio]{mshell}}]{Michael Shell}
	% or if you just want to reserve a space for a photo:
	
	%\begin{IEEEbiographynophoto}{Yanxiang Gong}
		%Yanxiang Gong is a master degree candidate with the School of Information and Communication Engineering at University of Electronic Science and Technology of China(UESTC). His research interests include pattern recognition, computer vision and deep learning.
	%\end{IEEEbiographynophoto}
	
	% if you will not have a photo at all:
	%\begin{IEEEbiographynophoto}{Linjie Deng}
	%Biography text here.
	%\end{IEEEbiographynophoto}
	
	% insert where needed to balance the two columns on the last page with
	% biographies
	%\newpage
	
	%\begin{IEEEbiographynophoto}{Mei Xie}
	%Biography text here.
	%\end{IEEEbiographynophoto}
	
	% You can push biographies down or up by placing
	% a \vfill before or after them. The appropriate
	% use of \vfill depends on what kind of text is
	% on the last page and whether or not the columns
	% are being equalized.
	
	%\vfill
	
	% Can be used to pull up biographies so that the bottom of the last one
	% is flush with the other column.
	%\enlargethispage{-5in}
	
	% that's all folks
\end{document}